\documentclass[10pt,conference,a4paper]{IEEEtran}

\usepackage{xcolor}
\usepackage{amsmath}
\usepackage{amssymb}
\usepackage{algorithm}
\usepackage{algorithmic}
\usepackage{multirow,hhline,graphicx,array}
\usepackage{comment}
\usepackage{graphicx}
\usepackage{caption}

\DeclareMathOperator*{\argmax}{arg\,max}

\begin{document}

\title{Complex-Object Visual Inspection via Multiple Lighting Configurations}

\author{\IEEEauthorblockN{Maya Aghaei,
Matteo Bustreo,
Pietro Morerio,
Nicol\`o Carissimi,
Alessio {Del Bue},
Vittorio Murino}

\IEEEauthorblockN{
Pattern Analysis and Computer Vision (PAVIS),
Istituto Italiano di Tecnologia (IIT), Genova, Italy\\}
Email: \{maya.aghaei, matteo.bustreo, pietro.moreiro,
nicolo.carissimi, alessio.delbue, vittorio.murino\}@iit.it}

\maketitle
\begin{abstract}

The design of an automatic visual inspection system is usually performed in two stages. While the first stage consists in selecting the most suitable hardware setup for highlighting most effectively the defects on the surface to be inspected, the second stage concerns the development of algorithmic solutions to exploit the potentials offered by the collected data. 

In this paper, first, we present a novel illumination setup embedding four illumination configurations to resemble diffused, dark-field, and front lighting techniques. Second, we analyze the contributions brought by deploying the proposed setup in training phase only - mimicking the scenario in which an already developed visual inspection system cannot be modified on the customer site - and in evaluation phase. Along with an exhaustive set of experiments, in this paper, we demonstrate the suitability of the proposed setup for effective illumination of complex-objects, defined as manufactured items with variable surface characteristics that cannot be determined a priori. Moreover, we discuss the importance of multiple light configurations availability during training and their natural boosting effect which, without the need to modify the system design in evaluation phase, lead to improvements in the overall system performance.


\end{abstract}
\IEEEpeerreviewmaketitle

\section{Introduction}
\label{sec:intro}

Quality control is a fundamental process in the manufacturing pipeline. Since the '80s, automatizing the quality control task has been offering potentials to overcome limitations of  manual inspection \cite{chin1982automated}. As a consequence, successful applications of automatic visual inspection have been emerging year after year, and nowadays inspection systems are being employed in a vast number of industries, from food \cite{wu2013colour} and fabrics \cite{li2016deformable} to railways \cite{shang2018detection} and reconstruction \cite{chen2017self}.

In this regard, a standard visual inspection hardware setup is typically composed of a digital camera, optics, and an illumination system. The hardware setup is usually coupled with a customised software that controls the acquisition, evaluates the captured images, and eventually takes decisions based on the evaluated results. Hence, hardware selection is a fundamental task in the design of an automatic visual inspection system and is essentially driven by the characteristics of the object to be inspected \cite{see2017role}.

\begin{figure}
    \centering
    \includegraphics[width=0.45\textwidth]{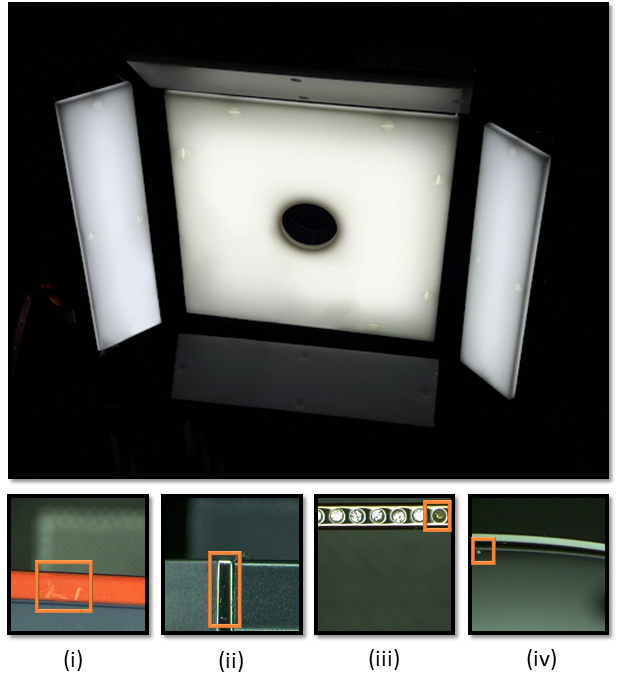}
    \caption{The proposed lighting system and four examples of defective images in our dataset as (i) worn-out paint, (ii) dots on the metal surface, (iii) missing decorations, and (iv) unexpected glass break. }
    \label{fig:defects}
    \vspace{-1em}
\end{figure}

For a given manufactured object, countless different models might exist in production having various material properties (specular, diffusive, directional, transparent) or geometrical shape (flat, curved, prismatic). The surface to be inspected might also contain patterns and adornments which should be distinguished from the undesirable irregularities (see Fig. \ref{fig:defects}). We define the object to be inspected, which is the subject of this work, as a \textit{complex-object} if its variable surface characteristics cannot be determined a priori, e.g. it can appear highly reflective and curved in one instance and opaque and prismatic in a different instance. This situation is not uncommon when inspecting assembled and/or decorated products, which can have custom finishing, based on customer requests.\footnote{Due to Non-Disclosure Agreement (NDA) restrictions in place, we cannot reveal the identity of the object inspected in this study.}

In this context, \textit{standard illumination techniques} comprising `front lighting', `back lighting', `diffuse lighting', `bright-field lighting' and `dark-field lighting' \cite{van1996choose} individually are not sufficient for this task as each of them is merely suitable for inspection of a few certain surface characteristics. Additionally, the surface attributes are not the only factor driving the choice of the illumination setup. In fact, \cite{martin2007practical} names \textit{immediate inspection environment} one of the three factors for an optimal lighting solution, and introduces object geometry and its support structure as two critical factors for the design of lighting solutions that may even limit the choice of standard illumination techniques. 

In this work, we aim to propose an illumination system which is capable of dealing with the challenges of automatic visual inspection of the complex-objects, and to define a methodology for analyzing the effect of the proposed illumination system on the final defect detection performance. In particular, we seek to study the impact of the proposed multi-lighting system when deployed in training phase only or in both training and evaluation phases. The first case is specifically relevant in the common situation where deployment of a novel acquisition system cannot be accomplished on the customer site, either due to industrial constraints or technical specifications. To summarize, our contributions are as follows:
\begin{itemize}
    \item We propose an acquisition setup composed of a multi-illumination system (diffused, dark-field and frontal illumination techniques) to guarantee high defect visibility (over 99\%, as reported by the annotators) on a wide selection of instances of the complex-object.
    \item We conduct exhaustive experiments to demonstrate the importance of the the multi-lighting system, even though merely deployed in training phase.
    \item We experimentally show that the multi-lighting setup deployment in the evaluation phase, when coupled with late-fusion of detections in each single-lighting conditions, leads to the highest  defect detection rate of the system.
\end{itemize}

\section{Related work}
\label{sec:related}

The list of successful applications of the visual inspection systems in the case of non-complex objects is long and in many cases the deployment of standard illumination techniques leads to significant improvements. For instance,  \cite{chang2016development} addresses \textit{touch panel glass} defect detection using dark-field illumination coupled with image processing techniques achieving $99\%$ accuracy on edge defect type, and an ad-hoc illumination technique such as injecting light beams perpendicularly in the glass achieves $100\%$ performance in scratch defect detection and its discrimination from dust \cite{ozturk2018real}.

Inspection of non-regular objects, however, has always been considered a challenging task where a combination of hardware and software techniques was required to achieve the desired outcome. For \textit{silver halide films} inspection, adopting a combination of dark-field illumination brought to the best results in detecting scratch and dust \cite{rufenacht2013automatic}. In certain inspection scenarios, such as small defects in \textit{automotive components}, standard illumination techniques were not found suitable. Thus, to ensure identifying defects when they are undetectable to the naked eyes, \cite{mery2017automatic} proposed to use x-ray imaging and achieved quality performance using SVM-linear classifier. Yet, in a similar case for detecting small defects on \textit{automobile casting aluminum parts}, deployment of x-ray imaging together with the most recent algorithms such as Feature Pyramid Networks leads to $0.51$ mAP in the best case scenario \cite{du2019approaches}.

In this paper, firstly we present a custom-designed illumination system comprising several heterogeneous lighting techniques including diffused, dark-field, and front lighting, under various camera exposure values to illuminate numerous defect types on a wide range of surface characteristics that a complex-object might be made of. In addition, we discuss that collecting data under various illumination configurations can be understood as representing an artifact in different modalities, although all the modalities are in practice offered in a single data format as the \textit{RGB image}. As also suggested in \cite{guo2015microscopy}, hereafter we will thus regard images acquired under different illumination configurations as different modalities.

Secondly, we provide exhaustive analysis on the potentials of the proposed system to be utilized in either training or evaluation phase. In many cases, the illumination system cannot be arbitrarily chosen or modified due to, for example, out of reach system specifications or cost related issues, especially in customer site (evaluation phase). Hence, we will investigate performance improvement brought by the developed system only in the data collection phase for training of algorithms only in provider site. Further, we experimentally demonstrate that mutual processing of multiple modalities in the form of late-fusion of single detections in each modality leads to considerable improvements in the performance of defect detection algorithms if employed in both training and evaluation phases, thus justifying the suitability of the proposed pipeline. 

In this regard, the work most similar to ours is the one proposed in \cite{park2016ambiguous}, where to detect and classify defects on a \textit{smartphone surface}, several images are taken with various cameras and light sources to ensure the visibility of defects in at least some of the images. However, differently from our proposal where images are taken with a single camera under varying illumination conditions, in \cite{park2016ambiguous}, as the images are taken with different cameras placed in different locations, the mutual processing of the collected images does not occur. 

Our motivations for proposing our design are threefold: first, in our proposal, only one camera is embedded, leading to a more cost-effective setup. Second, our proposed setup is designed to have a moderate physical weight enabling it to be carried e.g. with a robotic arm to spin around the complex-object and acquire images at different positions of it. Third, and possibly of more interest to the pattern recognition community, our proposed setup provides multiple instances of the same defective region that we empirically demonstrate to have a large improving impact on defect detection procedure.  

\section{Acquisition setup} 
\label{sec:acquisition}

\begin{figure}
    \centering
    \includegraphics[width=0.5\textwidth]{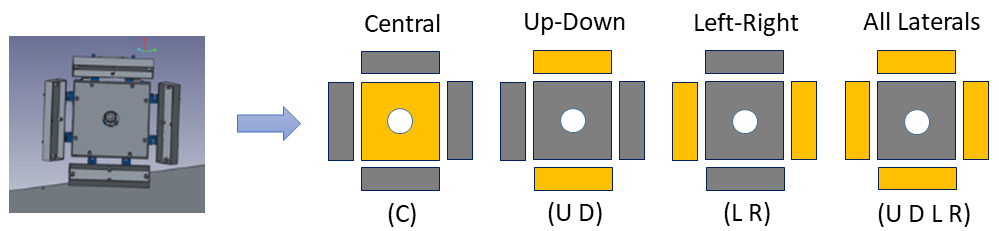}
    \caption{Illuminators in the proposed acquisition setup are set to activate and deactivate sequentially to resemble diffused, dark-field, and front lighting techniques within four illumination configurations (C, UD, LR, UDLR).}
    \label{fig:setup_lights}
    \vspace{-2em}
\end{figure}

\begin{figure}
    \centering
    \includegraphics[width=0.45\textwidth]{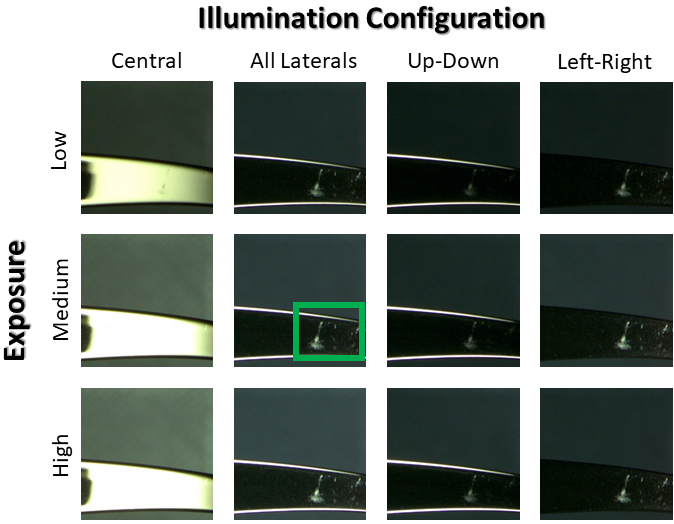}
    \caption{Images of a defective object in various illumination conditions. The labeled defect annotation by the annotator is shown with a green bounding-box.}
    \label{fig:modality}
    \vspace{-1em}
\end{figure}

Our proposed lighting setup is composed of five flat-dome lights that alternatively activate and deactivate in different combinations. The light positioning has been empirically studied such to reproduce diffused, dark-field and front lighting techniques, while producing the least possible glares on the specular surfaces. Our proposed setup can be seen in Fig. \ref{fig:setup_lights}. 

Dome light offers diffused, shadow-less, and uniform illumination even on shiny, curved, and uneven surfaces. In fact, flat-dome lights provide the same characteristics of dome lights, with the additional advantage of occupying less volume, as of the standard LED light. To minimize the reflectivity of the lighting system, which would make it visible when acquiring highly specular surfaces, we covered all the white flat-dome lights with dark collimator filters.

We identified four lighting configurations which allow the system to produce front lighting (Fig. \ref{fig:setup_lights}.C) and dark-field lighting in vertical (Fig. \ref{fig:setup_lights}.UD), horizontal (Fig. \ref{fig:setup_lights}.LR) and all lateral (Fig. \ref{fig:setup_lights}.UDLR) directions. Front lighting is mostly suitable for detecting color irregularities or flat defects, while dark-field lighting is extremely useful for acquiring effective images of defects related with surface irregularities such as scratches, bumps, or missing pieces. 

In addition to the four modalities and to ensure the appropriate illumination level of the acquired images of any surface independently from their reflective characteristics, each light configuration is activated for 3 different time lengths, mimicking 3 different camera shutter speeds (low, medium, high). Camera exposure time is set to be constant and longer than the maximum time of light activation. Trigger controls are configured such that lights and the camera are properly synchronized. In our study, all the images are acquired using a Basler acA2440-75uc camera and an Edmund Optics 16mm  F1.4 lens. The camera is placed at the ad-hoc hole presented in the center of the central light. In order to block out all the external environment light, the entire setup and the complex-object to be inspected were placed in a dark black box.

\section{Dataset}
\label{sec:dataset}
Given the described acquisition setup, the system can simultaneously acquire 12 images of the same object varying the illumination conditions (4 modalities, each with 3 exposures). A defect, depending on its type and the characteristics of the surface on which it appears, might be visible in all or only some of the $12$ captured images. For example, as in the case shown in Fig. \ref{fig:modality}, the defect is visible in all the images but images captured with central light with medium and high exposures. Note the significantly different representation each one of the light configurations offers from a single defect.

\begin{figure}
    \centering
    \includegraphics[width=0.5\textwidth]{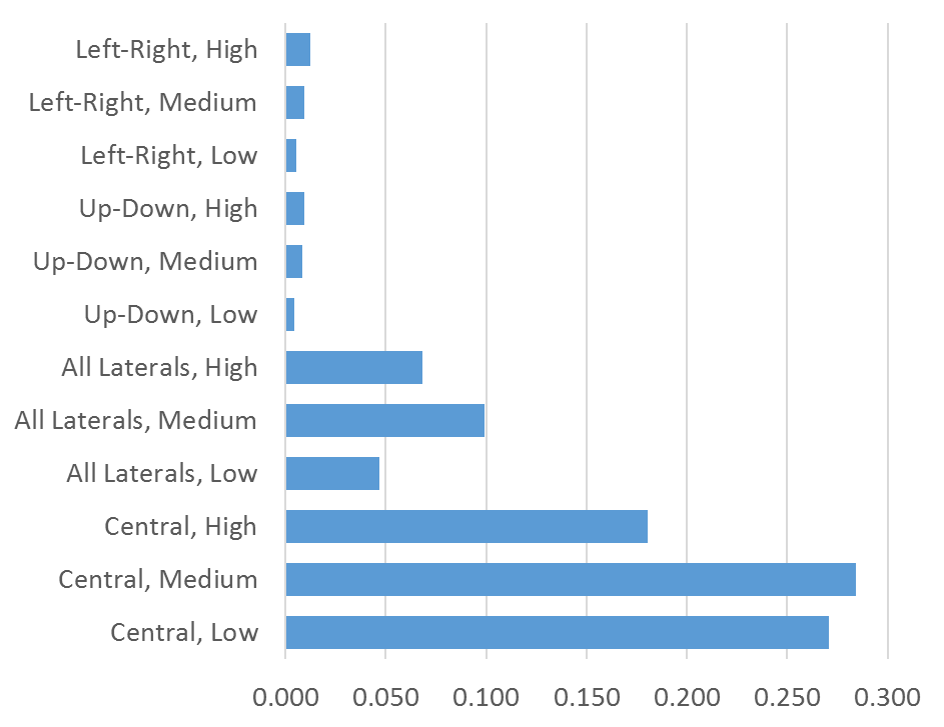}
    \caption{Normalized annotations frequency per illumination condition.}
    \label{fig:ann_freq}
    \vspace{-1em}
\end{figure}

Without predefined instructions on image choice, for each defective object, the annotators label the defect in only one of the images on which they can spot it, as shown by a green bounding-box in Fig. \ref{fig:modality}. Fig. \ref{fig:ann_freq} shows the normalized frequency of annotations for each illumination condition. We expand the single annotation on one image to all the 11 remaining images. If the existing defect on the object is not visible in any of the 12 images captured by the setup, the annotator indicates the non-visibility of the defect in the annotation tool. It is worth mentioning that, the developed setup enabled us to visualize and correctly annotate $99.2\%$ of the defects in a freely selected collection of complex-objects.

The collected dataset consists of $5,071$ defective regions of complex-objects, where each region may contain more than one defect. For each region, 12 images with varying illumination conditions are collected, obtaining a total number of $60,852$ images. For our experiments, we split the dataset object-wise in training, validation, and test set with the ratio of 70\%, 15\%, and 15\% respectively. 

\section{Methodology}
\label{sec:method}
Depending on the defect type and on the surface characteristics, the defect might be better visible in one or more than one image out of the 12 collected using the proposed setting. Given this, is selecting a conventional single illumination technique the most effective choice that the system provider can make? Can the system provider leverage the availability of the multi-modal data in training phase for improving uni-modal testing performance? Can different light conditions be considered as a natural data augmentation technique, or the resulting images are too correlated to actually bring any contribution during the model training? Can inspection scenarios benefit from the multi-modal data availability also in evaluation phase? In the following paragraphs, we explain our proposed methodology for responding to the aforementioned questions. 


\subsection{Study 1: Training and evaluation on one single modality}
\label{subsec:study1}
The most common situation when working with visual inspection systems consists of having the same illumination setup available in both training and evaluation phase, therefore it is fundamental to assess the best performing illumination modality. This scenario will be our baseline: only one illumination modality is available for training and evaluation. In this single modality scenario, we are interested in comparing the performances that can be obtained using each of different modalities, for better understanding the characteristics of our dataset and for exploring which light configuration may better help in solving our task.

Note that given the single modality scenario, only one quarter of the collected data is used, since the related images to all the other 3 modalities are discarded. Yet, in all the experiments, the selected light configuration includes all of its corresponding images taken under all the 3 exposures, unless stated differently.

\subsection{Study 2: Training on multiple modalities, evaluation on a single modality}
\label{subsec:study2}
As mentioned earlier, in some cases, visual inspection systems cannot be arbitrarily chosen or modified in evaluation phase. In this study, we aim to verify whether deploying a multi-modal inspection system only for acquiring images to be used for model training can lead to improved performances on the unmodified single modality evaluation setup.

In order to be comparable with the results of Study 1, we introduce images acquired using different illumination modalities keeping constant the number of images used during training. In other words, also in this experimental setup, only one quarter of the entire dataset is used. In this case, we choose two possible strategies to select dataset images to preserve:

\begin{itemize}
  \item Out of the 12 images available per each defective region, preserving 3 random images each from a different modality under one randomly selected exposure value only;
  \item Preserving only one quarter of the defective regions in the dataset, but using all of their 12 images acquired with all the light configurations and exposures.
\end{itemize}
Comparing the performances obtained by training the model on these two datasets will give us an insight on the comparative effectiveness of having either more defective objects or more modalities during training phase, given any of the single modalities in evaluation phase.

\subsection{Study 3: Training on all the images and modalities, test on a single modality}
\label{subsec:study3}
In Study 2 we discarded three quarter of the collected images for comparing the achieved results with the ones obtained in Study 1. Nevertheless, the proposed acquisition setup enables collecting 12 images per each object with no additional effort required for acquiring or annotating them in comparison to a single modality illumination system. 

The possibility of having a bigger training set to exploit, would raise expectations for modeling better the task to be solved. However, in complex-object defect detection scenario, it is not given that the additionally collected images, in fact, provide beneficial information for training a more effective model to be used in a single modality scenario. In case they do, it means that the system is able to transfer the information collected from one light modality to a different modality and that the system can better model the detection task even if only provided with modalities during training which are not available during evaluation. In this study, we aim to evaluate this hypothesis.

In comparison to Study 1 and Study 2, in Study 3 we are using the entire training set introduced in Sec. \ref{sec:dataset} which is four times bigger, while the test set remains intact.

\subsection{Study 4: Training and evaluation on multiple modalities}
\label{subsec:study4}
After having analyzed the impact of having a multi-modal lighting system available in training phase only, in this Study our aim is to verify the effectiveness of having the same multi-modal lighting system also in evaluation phase.

It is important to highlight that the images of the same defective region collected with different light illuminations share the same annotations and should produce the same output. Combining each generated output is, therefore, essential and we expect it can positively impact the final algorithm performance, as it has been shown in other scenarios \cite{aghaei2016multi}.

We propose the following fusing procedure: Let us define the set of 12 images of the same region collected varying the illumination conditions as $I = [i_1, i_2, \dots, i_{12}]$, let us define $B = [b_1, b_2, \dots, b_{M}]$ the set of the $M$ defective bounding-boxes detected in all $i_n$ images $\in I$, and let us also define  $C = [c_1, c_2, \dots, c_{M}]$ the set of the corresponding detection confidences given by the detection algorithm.
Our proposal is to apply Non-Maximal Suppression (NMS) algorithm over $B$ and replace on every $i_n \in I$ the output of the NMS algorithm. Given the NMS Intersection-over-Union (IoU) threshold as $\theta$, NMS algorithm operates as written in Algorithm \ref{alg:nms}.

\begin{algorithm}
\caption{Non-maximal Suppression}
\hspace*{\algorithmicindent} \textbf{Input} $B, C, \theta$\\
\hspace*{\algorithmicindent} \textbf{Initialization} $D \leftarrow \left \{  \right \}$
\begin{algorithmic}
\WHILE{$B \neq \emptyset$}
\STATE $\kappa \leftarrow \argmax C$
\STATE $K \leftarrow b_{\kappa}$
\STATE $B \leftarrow B - K$
\STATE $D \leftarrow D \cup K$
\FOR{$b_\zeta \in B$}
\IF{$IoU (K,b_\zeta) \geq \theta$}
\STATE $B \leftarrow B - b_\zeta$
\STATE $C \leftarrow C - c_\zeta$
\ENDIF
\ENDFOR
\ENDWHILE
\end{algorithmic}
\hspace*{\algorithmicindent} \textbf{Output} $D, C.$
\label{alg:nms}
\end{algorithm}

NMS operates in three steps: Firstly, it sorts all of the detected boxes based on their box confidence scores from high to low; secondly, it selects the box which has the highest box confidence score as the detection result; and finally, it discards other candidate boxes whose IoU value with the selected box is beyond the threshold. Within the remaining boxes, NMS repeats the above two steps until there is no remaining box in the candidate set $B$. 

In Study 4 we will compare the performances of the system when the model is trained on the entire multi-modal training set and evaluated on the entire test set, with and without applying the proposed late-fusion technique.

\section{Experimental setup, results and discussion}
\label{sec:exp-results}

In all the experiments discussed earlier in Sec.\ref{sec:method} for automatic defect detection, we used YOLO-v3 end-to-end detection pipeline \cite{redmon2018yolov3}, given its fast inference time and its ability to detect small defects.\footnote{ We would like to mention that a comparative study of detection algorithms is out of the scope of this paper.} 

YOLO-v3 detector has been originally trained over the COCO dataset \cite{lin2014microsoft}, then the weights of the network are adapted to our task using the transfer learning approach updating all the layers of the network. Training has been done on a NVIDIA GeForce RTX 2080 Ti GPU, with learning-rate = 0.0001, and momentum = 0.9.

As mentioned in Sec. \ref{sec:dataset}, the dataset is split into training, validation, and test sets. In the experiments where a subset of data is required (Study 1, 2 and 3), that subset is selected within training, validation, and test sets independently and the splits do not vary in the experiments belonging to the same Study, or shared among various Studies (for example, Test - C is common among Study 1, 2 and 3). This allows us to retain the comparability of the experiments from one Study to another. As in standard settings, the validation set is used to tune the parameters of the algorithm and the final results are reported on the test set. Each detection bounding-box proposed by the model is compared with the ground-truth and classified as:

\begin{itemize}
    \item True Positive (TP): the detection has IoU $\geq$ $threshold$ and it is therefore considered correct;
    \item False Positive (FP): the detection has IoU $<$ $threshold$ and it is therefore considered wrong;
    \item False Negative (FN): the ground-truth annotation has not been detected.
\end{itemize}

We report the results of all experiments using the standard metrics used in single-object (defect) detection as Precision, Recall, F1-score, and Average Precision (AP). Among the aforementioned metrics, Precision, Recall, and consequently, F1-score are reported after fixating the acceptance confidence threshold of the algorithm, in this work set to $0.7$. Precision is defined as $\frac{TP}{TP+FP}$, Recall as $\frac{TP}{TP+FN}$ and F1-score as $2\frac{Precision * Recall}{Precision + Recall}$. AP on the other side, summarizes the Precision-Recall curve as the weighted mean of Precision achieved at different confidence thresholds, with the increase in Recall from the previous threshold used as the weight and is calculated as $\sum_{t} (R_t - R_{t-1})P_t$, where $P_t$ and $R_t$ are the Precision and Recall at the $t$-th threshold.

To compare the results in the next sections, we will mainly refer to AP, since AP compared to F1-score considers Precision and Recall relations more globally \cite{boyd2012unachievable}. In this section, results are reported with a fixed $IoU=0.5$ threshold with the ground-truth among the experiment.

\subsection{Study 1: Training and evaluation on one single modality}

The results of the experiments discussed in Sec. \ref{subsec:study1} are given in Table \ref{table,1st}. The most effective configuration according to the AP is the one activating all the lateral lights to produce dark-field illumination from four directions. This configuration outperforms frontal light and dark-field illuminations in any of vertical and horizontal directions and it will be referred to as the baseline for the following studies.

\begin{table}[h!]
\centering
\caption{Results of Study 1}
\begin{tabular}{|c|c|c c c c|}
\hline 
Train                                                       & Test                                                        & Precision & Recall & F1-score & AP \\ \hline \hline
C                                                    & C                                                    & 63.53     & 45.84  & 53.25 & 29.97    \\ 
U D                                                     & U D                                                     & 61.69     & 44.95  & 52.01 & 29.11     \\ 
L R                                                  & L R                                                  & 58.56     & 41.07  & 48.28 & 25.52    \\ 
U D L R                                                     & U D L R                                                     & 61.06  & 52.73  & 56.82 & \textbf{34.69}    \\ \hline
\end{tabular}
\label{table,1st}
\end{table}


\subsection{Study 2: Training on multiple modalities, evaluation on a single modality} 

The results of the experiments discussed in Sec. \ref{subsec:study2} are reported in Table \ref{table,2nd}. Each training set has been generated 5 different random times for each experiment and results are given in $mean \pm std$ format in the AP column. Precision, Recall, and F1-score values are given for only the first trial. 

The results indicate, given the same number of images in the training set, maximizing the heterogeneity in the lighting modalities is more effective than acquiring more samples of defective objects with a limited set of illumination modalities.

\begin{table}[h!]
\centering
\caption{Results of Study 2}
\begin{tabular}{|c|c|c c c c|}
\hline
Train  ($N=5$)                                                       & Test                      & Precision & Recall & F1-score & AP \\ \hline \hline
\begin{tabular}[c]{@{}c@{}}All samples \\ 3 rand. modalities \end{tabular} & \multirow{2}{*}{C} & 66.28     & 38.00  & 48.30 & 25.74$\pm$2.75     \\ \cline{1-1} \cline{3-6} 
\begin{tabular}[c]{@{}c@{}}Quarter of samples\\ All modalities\end{tabular} &                           & 64.86     & 49.38  & 56.07 & 33.18$\pm$1.5  \\ \hline \hline
\begin{tabular}[c]{@{}c@{}}All samples \\ 3 rand. modalities \end{tabular} & \multirow{2}{*}{U D} & 66.98     & 36.94  & 47.62 & 27.17$\pm$3.9    \\ \cline{1-1} \cline{3-6} 
\begin{tabular}[c]{@{}c@{}}Quarter of samples\\ All modalities\end{tabular} &                           & 64.74     & 51.00  & 57.06 & 33.29$\pm$1.3     \\ \hline \hline
\begin{tabular}[c]{@{}c@{}}All samples \\ 3 rand. modalities \end{tabular} & \multirow{2}{*}{L R} & 68.48     & 36.90  & 47.96 & 27.45$\pm$2.9      \\ \cline{1-1} \cline{3-6} 
\begin{tabular}[c]{@{}c@{}}Quarter of samples\\ All modalities\end{tabular} &                           & 65.01     & 51.37   & 57.39 & \textbf{34.49$\pm$1.57}    \\ \hline \hline
\begin{tabular}[c]{@{}c@{}}All samples \\ 3 rand. modalities \end{tabular} & \multirow{2}{*}{U D L R} & 66.34     & 36.61   & 47.18 & 26.87$\pm$3.79     \\ \cline{1-1} \cline{3-6} 
\begin{tabular}[c]{@{}c@{}}Quarter of samples\\ All modalities\end{tabular} &                           & 63.53     & 48.32  & 54.89 & 31.84$\pm$0.65 \\ \hline 

\end{tabular}
\label{table,2nd}
\end{table}

Comparing the results of Study 2 with Study 1, it is noticeable that multi-modal training is beneficial for most of the single lighting modalities in evaluation and that the single-modal test performance is less dependent on the choice of the illumination modality if the algorithm is initially trained with multiple modalities.


\subsection{Study 3: Training on all the images and modalities, test on a single modality}

The results of Study 3 are listed in Table \ref{table,3rd}. Comparing these results with ones of Study 2, using a bigger training set leads to a considerable performance boost (at least $\sim18\%$). 
These results are a clear demonstration that acquiring more images using multiple light conditions is actually enriching the information provided to the model during training. Even in the case when 3 modalities out of 4 are not used in evaluation time, their availability during training makes the system able to better model the detection task to be solved, as it has been shown in other scenarios \cite{garcia2018modality}.

\begin{table}[h!]
\centering
\caption{Results of Study 3}
\begin{tabular}{|c|c|c c c c|}
\hline
Train                     & Test    & Precision & Recall & F1-score & AP \\ \hline \hline
\multirow{4}{*}{\rotatebox[origin=c]{90}{All Train}} & C       & 72.61     & 70.23  & 71.39  & 52.29     \\
                          & U D     & 70.69     & 71.22 & 70.95  & 52.27    \\ 
                          & L R     & 73.76     & 68.87   & 71.23 & \textbf{52.57}   \\ 
                          & U D L R & 72.11     & 70.37 & 71.23  & 52.38    \\ \hline
\end{tabular}
\label{table,3rd}
\end{table}

Eventually, it is worth noting that choosing any illumination modality to be used in production, after training the model with the multi-modal illumination system, would not bring significant variation in the detection performances.


\subsection{Study 4: Training and evaluation on multiple modalities}

\begin{table}[h!]
\centering
\caption{Results of Study 4}
\begin{tabular}{|c|c|c c c c|}
\hline
Train                                                               & Test     & Precision & Recall  & F1-score & AP  \\ \hline \hline
All Train                                                           & All Test & 72.26     & 70.18 & 71.20  & 52.08 \\ \hline
\begin{tabular}[c]{@{}c@{}}All Train\\ +\\ Late-fusion\end{tabular} & All Test  & 58.23     & 90.56  & 70.89 & \textbf{60.84}  \\ \hline
\end{tabular}
\label{table,4th}
\end{table}

The focus of the experiments until this point was given to the analysis of the effect of the presence of all or selected number of modalities in training while evaluation of the algorithms has been reported on single modalities. In Study 4, we aim to analyze whether it is possible to further improve the overall system performance having the availability of all the modalities also in evaluation phase. With this Study we can also assess the benefits which can be obtained with the deployment of our designed system in the operational scenario. The results of this study are reported in Table \ref{table,4th}.

\begin{figure}
    \hbox{\hspace{3em}\includegraphics[width=0.40\textwidth]{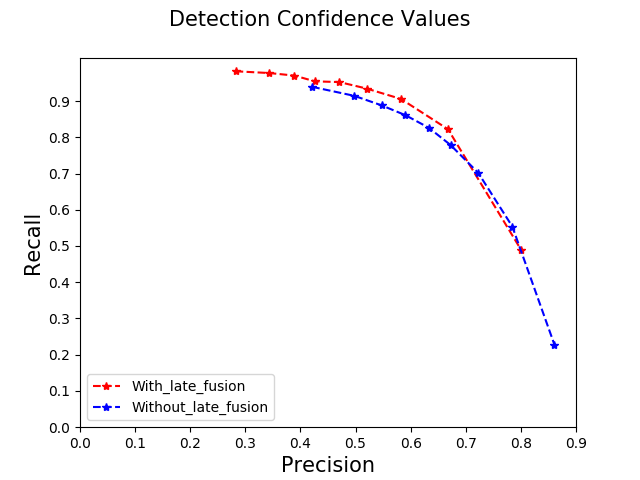}}
    \caption{Precision and Recall values \\ per confidence thresholds.}
    \label{fig:roc}
\end{figure}

\begin{figure*}
    \centering
    {\includegraphics[width=0.9\textwidth]{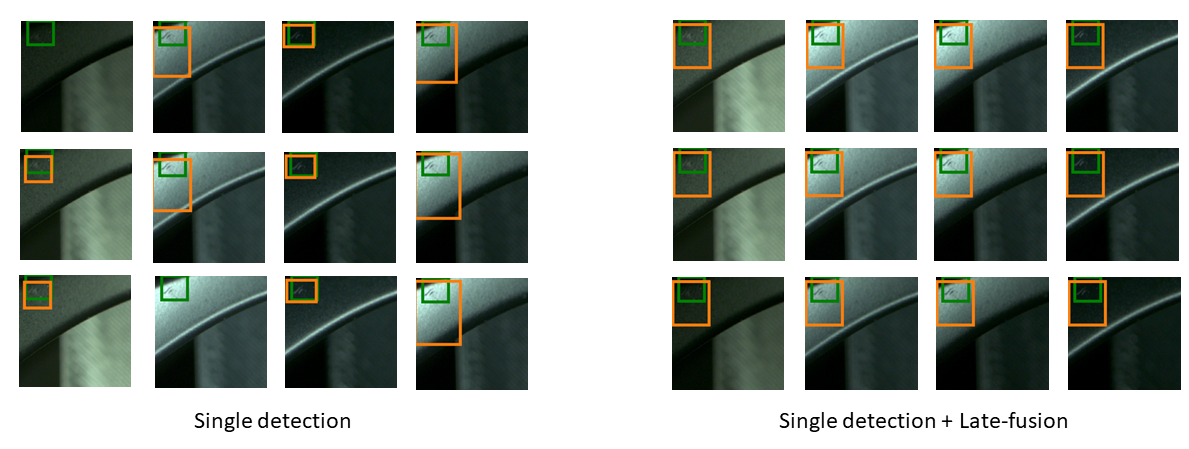}}
    \newline
    \centering
    {\includegraphics[width=0.9\textwidth]{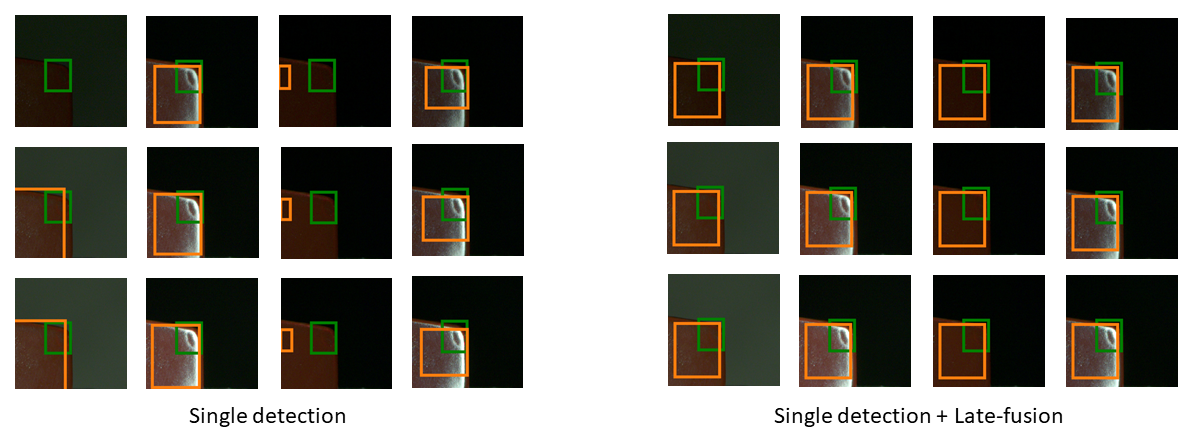}}
    \newline
    \centering
    {\includegraphics[width=0.9\textwidth]{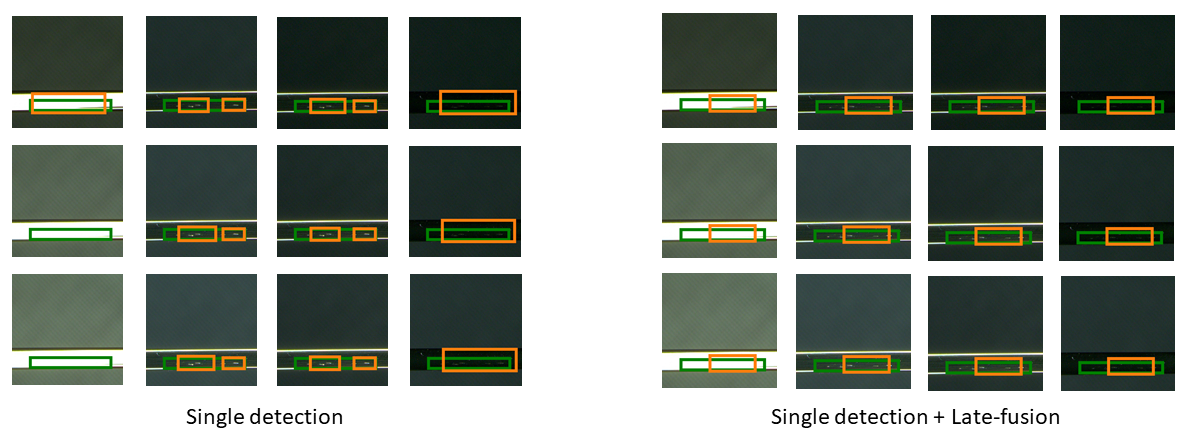}}
    \newline
    \caption{Three examples of successful qualitative results before and after applying the proposed late-fusion technique. Green bounding-boxes indicate the ground-truth while the orange indicates the final detection boundary. As it can be seen, applying the late-fusion of the results leads to a more coherent and correct final detection of defects in all the images of the defective object. Note how differently light configurations and exposures behave on different surface types and defects.}
\label{fig:quality}
\end{figure*}

Comparing the results given in Table \ref{table,4th} with ones in Table \ref{table,3rd}, having the availability of all the modalities in evaluation phase, leads to performance improvements only if the detection results obtained from each single illumination modality are properly combined, using the late-fusion technique proposed in Sec. \ref{subsec:study3}. Fig. \ref{fig:roc} shows the Precision and Recall values obtained at different detection confidence thresholds in  $\{0.1, 0.2, \dots, 0.9\}$, with and without employing the late-fusion technique. It can be observed that applying late-fusion leads to a higher Area Under Curve (AUC), thus higher AP. Employing late-fusion, Fig. \ref{fig:quality} shows three examples of the successful detections of defects employing late-fusion (on the right). The qualitatively better detections after applying late-fusion with regards to detections on single images can be appreciated in all the three cases.

\begin{figure*}
    \centering
    \includegraphics[width=0.8\textwidth]{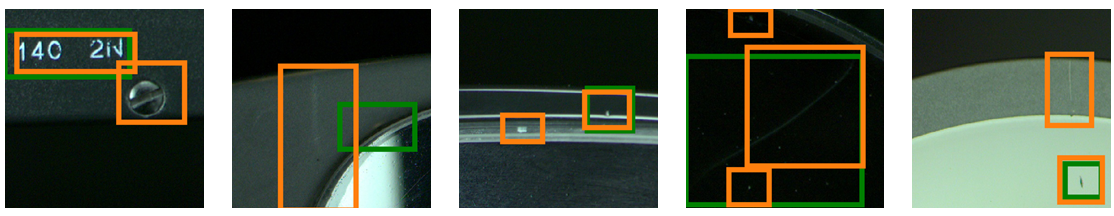}
    \caption{Five examples of missing detections or false alarms applying the proposed late-fusion. One can appreciate the visually similar-to-defect regions where false positives occur. In some of the images, the algorithm fails to detect the defect possibly due to the invisibility of the defect on the images.}
    \label{fig:failure}
    \vspace{-1em}
\end{figure*}

On the other side, Fig. \ref{fig:failure} shows five examples of failure cases even after the late-fusion technique in five defective images. In these cases, our observation is that the algorithm fails to detect a defect if it is not fairly visible in any of the images taken under any of the lighting conditions \cite{kokoschka1986visual}. Besides, false positive detections in some cases occur due to the presence of visually similar-to-defect artifacts on the images. This can be considered to confirm the importance of acquisition hardware setup design, and further, annotation process, for obtaining desirable results by the machine vision algorithms. In the cases where false positive detections are due to missing annotations, thus noise in the labels, the proposed method can be used to provide support for localization of defects to be fixed in the product revision departments in industries, or as an additional supervision method for further improvement of training procedure.

\section{Conclusion}

In this paper, we introduced our custom-designed acquisition setup for inspection of a complex-object, and discussed its suitability in visualizing a wide range of surface defects thanks to the proposed illumination setup which holds four standard illumination techniques comprised of diffused, dark-field, and front illumination in one place.

Further, we argued that deployment of the proposed setup might not be feasible in an inspection environment, thus we conducted four studies to exploit the role of each of the illumination sources and whether it is possible to exploit the potentials of the proposed setup when only deployed in training phase. The conclusions from the studies can be summarized as follows. In the case of deployment of the same single illumination modality in both training and evaluation phase, the most effective one is discovered to be activating all the lateral lights resembling dark-field illumination from four directions. However, given the same number of images in training set but with more modalities, the evaluation results on any of the single modalities are less dependent on the type of modality in evaluation phase. Nevertheless, exploiting more samples in all the modalities in training phase brings to a large improvement when evaluated on single modalities, justifying our proposed lighting setup to be employed at least for training purposes. The introduction of all the modalities in evaluation phase though does not lead to any substantial change with regards to a single modality illumination only, unless the proposed late-fusion technique is utilized which is when the highest performance of the proposed pipeline is achieved.

We believe our proposed acquisition setup and pattern analysis of the illumination modalities can be a source of intuition for other researchers in the industrial inspection field for the automatic examination of objects with highly complex characteristics.

\bibliography{2020_icpr_Lux_revision}

\begin{thebibliography}{10}

\bibitem{chin1982automated}
R.~T. Chin and C.~A. Harlow, ``Automated visual inspection: A survey,'' {\em
  IEEE Transactions on Pattern Analysis and Machine Intelligence}, no.~6,
  pp.~557--573, 1982.

\bibitem{wu2013colour}
D.~Wu and D.-W. Sun, ``Colour measurements by computer vision for food quality
  control--a review,'' {\em Trends in Food Science \& Technology}, vol.~29,
  no.~1, pp.~5--20, 2013.

\bibitem{li2016deformable}
Y.~Li, W.~Zhao, and J.~Pan, ``Deformable patterned fabric defect detection with
  fisher criterion-based deep learning,'' {\em IEEE Transactions on Automation
  Science and Engineering}, vol.~14, no.~2, pp.~1256--1264, 2016.

\bibitem{shang2018detection}
L.~Shang, Q.~Yang, J.~Wang, S.~Li, and W.~Lei, ``Detection of rail surface
  defects based on cnn image recognition and classification,'' in {\em
  International Conference on Advanced Communication Technology}, pp.~45--51,
  IEEE, 2018.

\bibitem{chen2017self}
J.-H. Chen, M.-C. Su, R.~Cao, S.-C. Hsu, and J.-C. Lu, ``A self organizing map
  optimization based image recognition and processing model for bridge crack
  inspection,'' {\em Automation in Construction}, vol.~73, pp.~58--66, 2017.

\bibitem{see2017role}
J.~E. See, C.~G. Drury, A.~Speed, A.~Williams, and N.~Khalandi, ``The role of
  visual inspection in the 21st century,'' in {\em Proceedings of the Human
  Factors and Ergonomics Society Annual Meeting}, vol.~61, pp.~262--266, SAGE
  Publications Sage CA: Los Angeles, CA, 2017.

\bibitem{van1996choose}
C.~H. Van~Dommelen, ``Choose the right lightning for inspection,'' {\em Test
  and Measurement World}, vol.~16, pp.~53--60, 1996.

\bibitem{martin2007practical}
D.~Martin, ``A practical guide to machine vision lighting,'' {\em Midwest Sales
  and Support Manager}, pp.~1--3, 2007.

\bibitem{chang2016development}
M.~Chang, B.-C. Chen, J.~L. Gabayno, and M.-F. Chen, ``Development of an
  optical inspection platform for surface defect detection in touch panel
  glass,'' {\em International Journal of Optomechatronics}, vol.~10, no.~2,
  pp.~63--72, 2016.

\bibitem{ozturk2018real}
{\c{S}}.~{\"O}zt{\"u}rk and B.~Akdemir, ``Real-time product quality control
  system using optimized gabor filter bank,'' {\em The International Journal of
  Advanced Manufacturing Technology}, vol.~96, no.~1-4, pp.~11--19, 2018.

\bibitem{rufenacht2013automatic}
D.~R{\"u}fenacht, G.~Trumpy, R.~Gschwind, and S.~S{\"u}sstrunk, ``Automatic
  detection of dust and scratches in silver halide film using polarized
  dark-field illumination,'' in {\em IEEE International Conference on Image
  Processing}, pp.~2096--2100, IEEE, 2013.

\bibitem{mery2017automatic}
D.~Mery and C.~Arteta, ``Automatic defect recognition in x-ray testing using
  computer vision,'' in {\em IEEE Winter Conference on Applications of Computer
  Vision}, pp.~1026--1035, IEEE, 2017.

\bibitem{du2019approaches}
W.~Du, H.~Shen, J.~Fu, G.~Zhang, and Q.~He, ``Approaches for improvement of the
  x-ray image defect detection of automobile casting aluminum parts based on
  deep learning,'' {\em NDT \& E International}, vol.~107, p.~102144, 2019.

\bibitem{guo2015microscopy}
K.~Guo, Z.~Bian, S.~Dong, P.~Nanda, Y.~M. Wang, and G.~Zheng, ``Microscopy
  illumination engineering using a low-cost liquid crystal display,'' {\em
  Biomedical optics express}, vol.~6, no.~2, pp.~574--579, 2015.

\bibitem{park2016ambiguous}
Y.~Park and I.~S. Kweon, ``Ambiguous surface defect image classification of
  amoled displays in smartphones,'' {\em IEEE Transactions on Industrial
  Informatics}, vol.~12, no.~2, pp.~597--607, 2016.

\bibitem{aghaei2016multi}
M.~Aghaei, M.~Dimiccoli, and P.~Radeva, ``Multi-face tracking by extended
  bag-of-tracklets in egocentric photo-streams,'' {\em Computer Vision and
  Image Understanding}, vol.~149, pp.~146--156, 2016.

\bibitem{redmon2018yolov3}
J.~Redmon and A.~Farhadi, ``Yolov3: An incremental improvement,'' {\em arXiv
  preprint arXiv:1804.02767}, 2018.

\bibitem{lin2014microsoft}
T.-Y. Lin, M.~Maire, S.~Belongie, J.~Hays, P.~Perona, D.~Ramanan,
  P.~Doll{\'a}r, and C.~L. Zitnick, ``Microsoft coco: Common objects in
  context,'' in {\em European Conference on Computer Vision}, pp.~740--755,
  Springer, 2014.

\bibitem{boyd2012unachievable}
K.~Boyd, V.~S. Costa, J.~Davis, and C.~D. Page, ``Unachievable region in
  precision-recall space and its effect on empirical evaluation,'' in {\em
  Proceedings of the International Conference on Machine Learning.}, p.~349,
  NIH Public Access, 2012.

\bibitem{garcia2018modality}
N.~C. Garcia, P.~Morerio, and V.~Murino, ``Modality distillation with multiple
  stream networks for action recognition,'' in {\em Proceedings of the European
  Conference on Computer Vision}, pp.~103--118, 2018.

\bibitem{kokoschka1986visual}
S.~Kokoschka and H.~Bodmann, ``Visual inspection of sealing rings—a case
  study on lighting and visibility,'' {\em Lighting Research \& Technology},
  vol.~18, no.~2, pp.~98--101, 1986.

\end{thebibliography}
\bibliographystyle{ieeetr}

\end{document}